\pdfoutput=1

\documentclass[11pt]{article}

\usepackage{acl}

\usepackage{times}
\usepackage{latexsym}

\usepackage[T1]{fontenc}

\usepackage[utf8]{inputenc}
\usepackage{times}
\usepackage{latexsym}
\usepackage{url}
\usepackage{enumitem}
\usepackage{amsmath,amssymb,amsfonts}
\usepackage{algorithmic}
\usepackage{caption}
\usepackage{subcaption}
\usepackage{graphicx}
\usepackage{textcomp}
\usepackage{url}
\usepackage{makecell}
\usepackage{multirow}
\usepackage{booktabs}
\usepackage{array}
\usepackage{microtype}
\usepackage{colortbl}

\usepackage{multirow}
\usepackage{booktabs}
\usepackage{hyperref}

\usepackage{hyperref}       
\usepackage{amsfonts}       
\usepackage{nicefrac}       
\usepackage{xcolor}         
\usepackage{tcolorbox}

\newcommand{\thickhline}{%
    \noalign {\ifnum 0=`}\fi \hrule height 1pt
    \futurelet \reserved@a \@xhline
}

\usepackage{comment}
\usepackage{todonotes}
\usepackage{xcolor}

\usepackage{microtype}
\usepackage[cjk]{kotex}
\newcommand\blfootnote[1]{%
  \begingroup
  \renewcommand\thefootnote{}\footnote{#1}%
  \addtocounter{footnote}{-1}%
  \endgroup
}

\definecolor{lp}{HTML}{CBC3E3}

\title{Open Ko-LLM Leaderboard: Evaluating Large Language Models in Korean with Ko-H5 Benchmark}

\author{Chanjun Park, Hyeonwoo Kim, Dahyun Kim, Seonghwan Cho \\ {\bf \large Sanghoon Kim, Sukyung Lee, Yungi Kim, Hwalsuk Lee$^{\dagger}$}\\
\\
  Upstage AI \\
  \texttt{\{chanjun.park, choco\_9966, kdahyun, hwalsuk.lee\}@upstage.ai}}

\begin{document}
\maketitle

\begin{abstract}
\blfootnote{$^\dagger$ Corresponding Author }
This paper introduces the Open Ko-LLM Leaderboard\footnote{\url{https://huggingface.co/spaces/upstage/open-ko-llm-leaderboard}} and the Ko-H5 Benchmark as vital tools for evaluating Large Language Models (LLMs) in Korean. Incorporating private test sets while mirroring the English Open LLM Leaderboard, we establish a robust evaluation framework that has been well integrated in the Korean LLM community. We perform data leakage analysis that shows the benefit of private test sets along with a correlation study within the Ko-H5 benchmark and temporal analyses of the Ko-H5 score. Moreover, we present empirical support for the need to expand beyond set benchmarks. We hope the Open Ko-LLM Leaderboard sets precedent for expanding LLM evaluation to foster more linguistic diversity.
\end{abstract}

\section{Introduction}
The emergence of Large Language Models (LLMs)~\cite{zhao2023survey} have also introduced an ever growing demand for robust evaluation frameworks for LLMs. While multiple benchmarks~\cite{open-llm-leaderboard, alpaca_eval, zheng2023judging, 2023opencompass} have been proposed for a more holistic evaluation of LLMs, they are mostly limited to the English language.
Recognizing the need to expand the mostly English-centric LLM benchmarks to other languages such as Korean, we introduce the ``Open Ko-LLM Leaderboard'' and the ``Ko-H5 Benchmark''.

The Open Ko-LLM Leaderboard is built on the following two principles: i) alignment with the English Open LLM Leaderboard~\cite{open-llm-leaderboard} and ii) private test sets.
Enabling straightforward comparison between the two leaderboard results, following the well-established composition of the Open LLM Leaderboard is key to the successful integration of the Open Ko-LLM Leaderboard in the Korean LLM community. Further, our private test sets allow for robust evaluation of a plethora of models in the wild without significant worry of data contamination on the tested benchmarks~\cite{sainz2023nlp,zhou2023don,balloccu2024leak}. We show that our private test sets have little overlap with some of the most popular training datasets used by top models in the Open Ko-LLM Leaderboard, empirically solidifying the argument for private test sets.

To reveal various key insights, we perform an extensive multi-faceted analysis.
For instance, correlation between the tasks that constitute the Ko-H5 benchmark shows that the newly added dataset, {\it i.e.}, Ko-CommonGen v2, differentiates the Open Ko-LLM Leaderboard from the English Open LLM Leaderboard by bringing more diversity to the evaluation suite.
Additionally, analysis of the improvements in the Ko-H5 score over time for differently sized models presents insights into a potential critical model size that enables rapid performance improvement.
Another temporal analysis of the Ko-H5 benchmark scores with respect to various model types brings \textit{quantitative} support for the notion that improvements in pretrained models lead to improvements in instruction-tuned models. 
Further analysis reveals a relatively quick saturation of certain task scores, indicating the need to move beyond a set benchmark. In other words, a shift towards a more holistic evaluation scheme that better adheres to real-world use-cases is needed. Building on the analytical results on score changes for each task of the top performing models, we offer a practical criteria of judging when to expand the evaluation suite for LLMs.

Our contributions can be summarized as follows:
\begin{itemize}
    \item We introduce the ``Open Ko-LLM Leaderboard'' and ``Ko-H5 Benchmark'' for expanding robust and widespread evaluation of Korean LLMs.
    \item We address the issue of data contamination by using private test sets for fair model evaluation, ensuring minimal overlap with popular training datasets.
    \item We present several analyses that highlight diverse insights ranging from inter-benchmark correlation to change of the benchmark scores over time, aggregated by model size and type and individual tasks.
    \item We offer practical criteria of when to expand beyond a set benchmark, emphasizing the need for diverse tasks to continually enhance LLM evaluation.
\end{itemize}

\section{Related Work and Background}\label{sec:related_work}
\subsection{LLM Leaderboard}
In the rapidly evolving landscape of Large Language Models (LLMs), evaluation of model performance from various aspects has become crucial. This is facilitated by various leaderboards, each designed to benchmark specific aspects of LLM capabilities. Among them, the Open LLM Leaderboard~\cite{open-llm-leaderboard} is prominent, operated by Hugging Face, a leading machine learning platform~\cite{jain2022hugging}. It provides a global benchmark for LLMs developed by many companies and research institutions. The leaderboard assesses models across six diverse tasks, including the AI2 Reasoning Challenge (ARC, in short)~\cite{clark2018think} for science questions, HellaSwag~\cite{zellers2019hellaswag} for commonsense inference, Massive Multitask Language Understanding (MMLU, in short)~\cite{hendrycks2020measuring} for natural language understanding ability, TruthfulQA~\cite{lin2021truthfulqa} for evaluating truthfulness, Winogrande~\cite{sakaguchi2021winogrande} for commonsense reasoning, and GSM8k~\cite{cobbe2021training} for mathematical reasoning problems.

AlpacaEval Leaderboard~\cite{alpaca_eval}, HELM Leaderboard~\cite{lee2023holistic}, and Hallucinations Leaderboard~\cite{HughesBae2023} each offer unique perspectives on model evaluation. The AlpacaEval Leaderboard evaluates the instruction following abilities of LLMs in a variety of natural language tasks, while HELM provides a holistic framework for evaluating LLMs across various scenarios. The Hallucinations Leaderboard specifically targets the phenomenon of hallucinations in outputs of LLMs, using benchmarks like TruthfulQA and HaluEvals~\cite{li2023halueval}.

For developers focused on code generation, the Big Code Models Leaderboard~\cite{bigcodeleaderboard} provides a competitive space to evaluate models using the HumanEval benchmark and MultiPL-E~\cite{cassano2022multipl}, emphasizing the multilingual capabilities of code-generating LLMs. The Open ASR Leaderboard~\cite{open-asr-leaderboard} assesses the evaluation of automatic speech recognition models, using metrics such as Word Error Rate and Real-Time Factor. The LLM Perf Leaderboard~\cite{llm-perf-leaderboard} dives into the computational aspects, assessing LLMs across different hardware, backends, and optimization settings, focusing on latency, throughput, memory, and energy efficiency.

\subsection{Korean LLM Leaderboard}
\begin{figure*}[t!]
    \centering
    \resizebox{0.80\linewidth}{!}{
\includegraphics{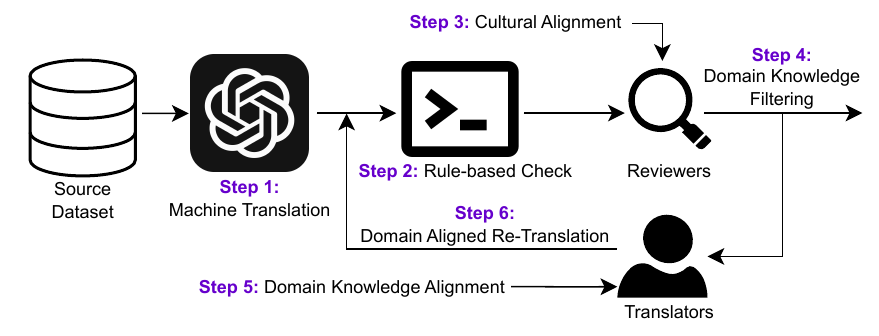}
    }
    \caption{Data curation process for the Ko-H5 benchmark. We perform thorough human review of the machine translation results by culturally aligning the reviewers with the Korean language. Additionally, we perform filtering for data that require specific domain knowledge and re-translate them with translators that are trained with the required domain knowledge.}
    \label{fig:translate}
\end{figure*}

Historically, the development of benchmarks and leaderboards for LLMs has been heavily skewed towards English~\cite{naveed2023comprehensive}, resulting in a rich array of evaluation benchmarks and platforms for English language models. Notable examples include GLUE~\cite{wang2018glue}, SuperGLUE~\cite{wang2019superglue}, and the aforementioned leaderboards. They have significantly advanced the field by providing standardized and diverse evaluation metrics. However, their focus on English has limited their applicability to other languages, especially those with unique linguistic characteristics like Korean.

Meanwhile, the research and development in evaluation of Korean LLMs have been markedly sparse. This is because the Korean language presents unique challenges for the evaluation of LLMs due to its distinct syntax and semantics~\cite{park2020empirical}.
This scarcity leads to a significant opportunity for the development of Korean LLMs evaluation landscape. To the best of our knowledge, the ``Open Ko-LLM Leaderboard'' is the first effort to offer a comprehensive and tailored evaluation platform for Korean LLMs.
Our initiative is not merely an extension of existing leaderboard to a new language; it is an endeavor to establish a foundation for the Korean LLMs evaluation ecosystem. This involves developing new benchmarks and metrics that are specifically designed to assess the nuances of the Korean language. We believe that our efforts will help the global advancement of AI by bringing more linguistic diversity to the evaluation of LLMs.

\section{Ko-H5 and Open Ko-LLM Leaderboard}
\subsection{Motivation}
As discussed in Section~\ref{sec:related_work}, many benchmarks for the evaluation of LLMs have particularly focused on the English language~\cite{chang2023survey}. Subsequently, benchmarks for other languages are trailing behind substantially~\cite{magueresse2020low,ranathunga2023neural}.
However, establishing benchmarks for other languages is very challenging, as it requires an understanding of the structural and characteristic differences of those languages. Meanwhile, this endeavor becomes paramount for a more global and linguistic diverse adaptation of the LLMs.
Recognizing the above, we have built the Open Ko-LLM Leaderboard along with its Ko-H5 benchmark as a significant first step towards the evaluation of open-source LLMs in the Korean language.
In doing so, we adhere to the following key principles:
\begin{itemize}
    \item {\bf Alignment with the Open LLM Leaderboard}: To facilitate direct comparison of advancements on the Open Ko-LLM Leaderboard with those on the global Open LLM Leaderboard, we have aligned our leaderboard accordingly.
    \item {\bf Private test sets}: To enable robust comparison of a wide range of models in the wild with little fear of data contamination, we adhere to the use of private test sets.
\end{itemize}
In this paper, we suggest the above two principles as a solid foundation for extending the evaluation of LLMs to other languages as well.

\subsection{Ko-H5}
\begin{table}[t!]
    \centering
    \resizebox{1.0\linewidth}{!}{
        \begin{tabular}{lcc}
        \toprule
        Dataset & \# Samples & License \\
        \hline
        Ko-ARC & 1.1K & CC-BY-SA \\ 
        Ko-HellaSwag & 10.0K & MIT \\
        Ko-MMLU & 14.0K & CC-BY-SA \\
        Ko-TruthfulQA & 0.8K & Apache license 2.0 \\
        Ko-CommonGen v2 & 0.8K & Apache license 2.0 \\ 
        \bottomrule
        \end{tabular}
        }
    \caption{Number of samples and license information for each of the datasets in the Ko-H5 benchmark.}
    \label{tab:dataset_samples}
\end{table}

\begin{table*}[t!]
\centering
\resizebox{0.9\linewidth}{!}{
\begin{tabular}{cccccc}
\toprule
~&Ko-ARC & Ko-HellaSwag & Ko-MMLU & Ko-TruthfulQA & Ko-CommonGen v2 \\ \midrule
KoUltrafeedback~\footnote{\url{https://huggingface.co/datasets/maywell/ko_Ultrafeedback_binarized}} & 0.24\% & 0.78\%       & 0.92\%  & 0.02\%        & 0.10\%          \\
KoOpenOrcaPlatypus~\footnote{\url{https://huggingface.co/datasets/kyujinpy/KOR-OpenOrca-Platypus-v3}} & 0.18\% & 0.63\%       & 0.82\%  & 0.03\%        & 0.10\%          \\
KoAlpaca~\footnote{\url{https://huggingface.co/datasets/beomi/KoAlpaca-v1.1a}} & 0.19\% & 0.48\%       & 0.55\%  & 0.02\%        & 0.06\%          \\ 
\bottomrule 
\end{tabular}
}
\caption{Overlap percentage of the Ko-H5 private test sets with popular training data used by top performing models in the Open Ko-LLM Leaderboard. After performing both exact deduplication and minhash deduplication on each of the training and test datasets, we paired each training data with test sets and conducted minhash deduplication again on these joined pairs. Note that we performed a very aggressive deduplication with a similarity threshold of 0.05, an n-gram size of 20, and a minimum length of 30, 20, 10, 10, 30 for Ko-ARC, Ko-HellaSwag, Ko-MMLU, Ko-TruthfulQA, and Ko-CommonGen v2 respectively. Despite the aggressive deduplication, the overlap percentage are all under one percent, sometimes by substantial margins, showing how private test sets prevent data contamination.}
\label{tab:private_test}
\end{table*}

\paragraph{Curation process.}
The Ko-H5 benchmark is composed of multiple datasets, some of which are derived from the original English datasets used in the Open LLM Leaderboard, while some are built from scratch.

First, Korean ARC~\cite{clark2018think}, Hellaswag~\cite{zellers2019hellaswag}, Truthful QA~\cite{lin2021truthfulqa}, and MMLU~\cite{hendrycks2020measuring} are derived from their counterparts via thorough machine and human translation process, as illustrated in Figure~\ref{fig:translate}.
To better ensure cultural and linguistic relevance of the derived datasets to Korean, we have undertaken a rigorous human review process, where a total of 35 translation review experts conducted the review. The review cost amounted to a total of $80,000$ USD for Ko-ARC, Ko-MMLU, and Ko-TruthfulQA, while Ko-HellaSwag did not undergo manual review since its large size requires a high estimated cost of $720,000$ USD. Detailed information about the professional translation reviewers can be found in Appendix~\ref{appn:A}, and their workspace interface is presented in Appendix~\ref{appn:B}. 

Specifically, we first translate the source datasets by utilizing GPT-4, with the prompts shown in Appendix~\ref{appn:D}, for scalable translation.
Then, a rule-based check~\cite{costa2022no} is performed to detect simple translation errors. Thereafter, reviewers are reinforced with cultural alignment of the Korean language before conducting manual review. The reviewed translation results are then filtered based on whether they require specific domain knowledge or not.
As some of source datasets contain data that require domain specific knowledge such as maths and science, the above step is paramount in obtaining a well-curated benchmark dataset in the Korean language.
 An example of such data can be found in Figure~\ref{fig:mmlu_trans} in Appendix~\ref{appn:D}.
The filtered data in the aforementioned step are sent to translators who are proficient in the specific domain knowledge via the domain knowledge alignment step.
Lastly, a domain aligned re-translation of the filtered data is performed and the results are sent back to the rule-based check step.

Different from the above, the Korean CommonGen v2 is curated from scratch, inspired by CommonGen~\cite{lin2019commongen}. 
The Ko-CommonGen v2 task is mainly aimed at testing models on generating common knowledge. 
Note that Ko-CommonGen v2 brings more diversity to the Ko-H5 benchmark (see Sec.~\ref{sec:corr} for empirical evidence) and differentiates the Open Ko-LLM Leaderboard from its English counterpart.

\paragraph{Dataset sizes.}
The sizes and licenses of each dataset in the Ko-H5 benchmark are detailed in Table~\ref{tab:dataset_samples}. The licenses listed in the Table~\ref{tab:dataset_samples} are derived from the original English datasets when possible, all of which are free for redistribution. In the case of Ko-MMLU and Ko-HellaSwag, they are composed of more than 10K evaluation sets, a relatively large compared to other datasets. On the other hand, Ko-ARC, Ko-TruthfulQA, and Ko-CommonGen v2 are comprised of approximately 1,000 evaluation data each.

These differences reflect the characteristics of each dataset. For instance, Ko-MMLU and Ko-HellaSwag necessitate a larger samples to broadly assess various natural language understanding abilities and commonsense reasoning capabilities. Conversely, Ko-ARC, Ko-TruthfulQA, and Ko-CommonGen v2 focus on more specialized abilities such as domain-specific knowledge, truthfulness, and common sense generation, respectively, where a smaller number of high-quality samples may be more appropriate for evaluation.

\subsection{Open Ko-LLM Leaderboard}
The Open Ko-LLM Leaderboard represents a landmark development in the evaluation of Korean language models, meticulously replicating the framework established by Open LLM Leaderboard of Hugging Face~\cite{wolf2019huggingface}. This strategic decision to adopt the same platform reflects our commitment to maintaining a standardized, high-quality benchmarking system.
In doing so, researchers and developers familiar with the Open LLM Leaderboard can seamlessly transition to engaging with the Open Ko-LLM Leaderboard, fostering greater participation and collaboration in the development of Korean LLMs.

\section{Empirical Analysis}

\subsection{Private Test Set Overlap with Popular Training Datasets}
One of the key elements of the Ko-H5 benchmark is the private nature of the test sets.
By keeping the benchmark datasets private, we ensure robust and fair evaluation of LLMs with minimal data leakage. Note that while the original datasets in the H4 benchmark may face data leakage issues~\cite{deng2023benchmark} due to their public availability, our Ko-H5 benchmark datasets are kept private after being meticulously curated by human experts.

For analytical purposes, we select some of the most popular training datasets used by top performing models in the Open Ko-LLM Leaderboard and perform a data leakage study with the Ko-H5 benchmark datasets. First, deduplication on each of the training datasets and the Ko-H5 benchmark datasets is performed independently to remove any potential overlap inherent in each of the datasets. Then, the training datasets and the benchmark datasets are pairwise combined, where the combined datasets are also deduplicated. We summarize the percentage of the data samples that are removed from the Ko-H5 benchmark datasets in the aforementioned deduplication process in Table~\ref{tab:private_test}.

As seen from the table, there is little overlap of the benchmark datasets with some of the most popular training data used for developing Korean LLMs.
Specifically, even the highest overlap percentage is less than one percent for Ko-MMLU and KoUltrafeedback. 
Given the aggressive setting of the parameters such as the similarity threshold, the above results highlight the fact that private test sets substantially reduce data leakage risks in open evaluation benchmarks for LLMs.

\subsection{Correlation Within the Ko-H5 Benchmark}
\label{sec:corr}

\begin{figure}[t!]
    \centering
    \resizebox{0.99\linewidth}{!}{
\includegraphics{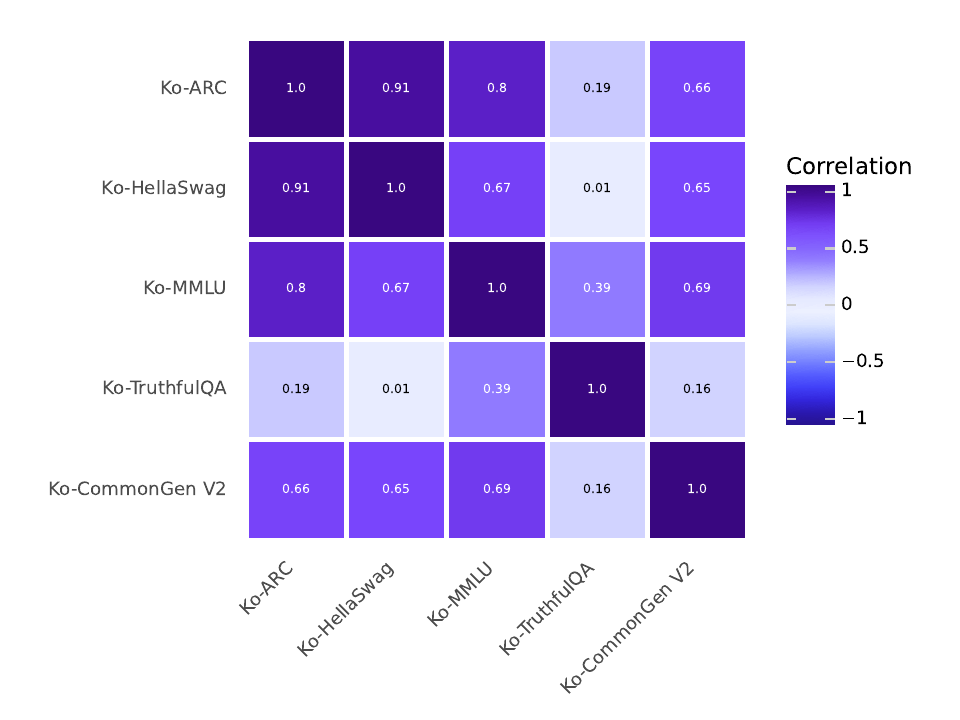}
    }
    \caption{Correlation between the different tasks in the Ko-H5 benchmark are shown in a heatmap format, with values ranging from $-1$ to $1$. Generally speaking, Ko-TruthfulQA and Ko-CommonGen v2 have lower correlation with other tasks.}
    \label{fig:correlation}
\end{figure}

\begin{figure*}[t!]
    \centering
    \resizebox{1.00\linewidth}{!}{
\includegraphics{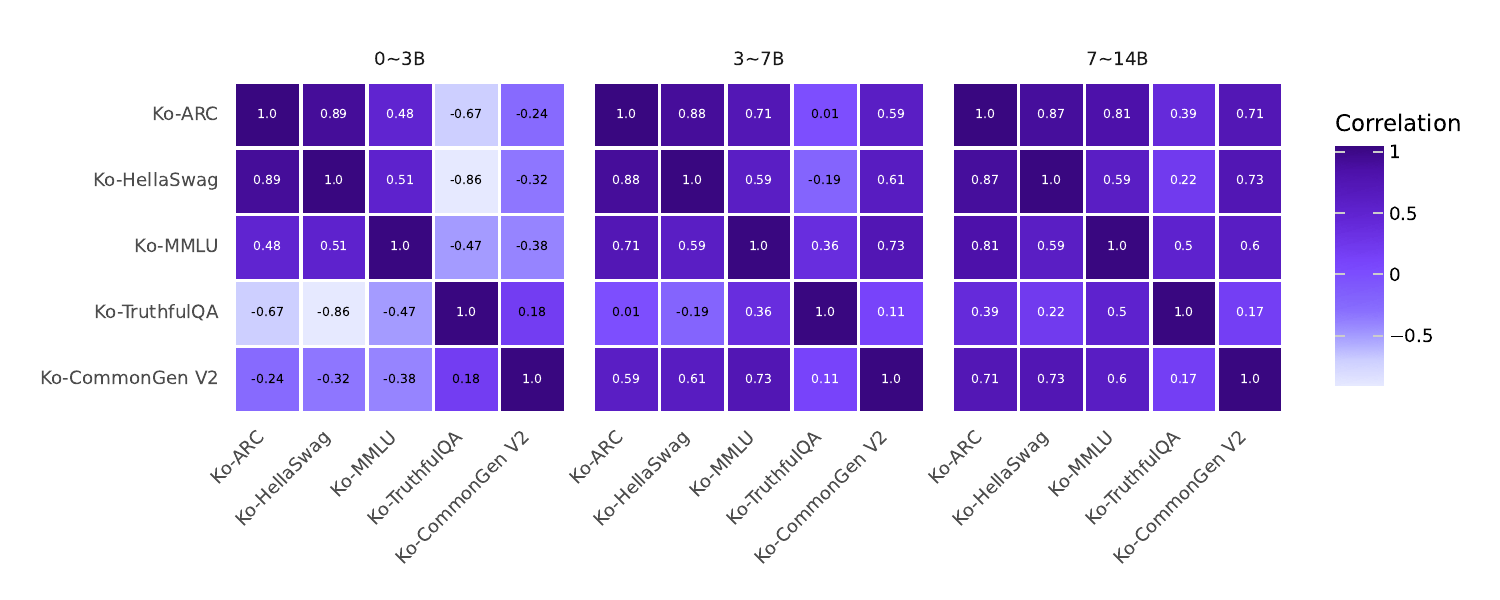}
    }
    \caption{Correlation between the different tasks in the Ko-H5 benchmark for different model size brackets are shown. The overall trend changes noticeably as the model size increases. Specifically, Ko-TruthfulQA and Ko-CommonGen v2 show low, or sometimes negative, values with other tasks in smaller model sizes whereas bigger models report higher correlation values.}
    \label{fig:correlation_per_size}
\end{figure*}

We perform a correlation study between the Ko-H5 benchmark datasets.
In particular, we focus on the correlation of the Ko-CommonGen v2 with the other benchmark datasets as the Ko-CommonGen v2 was newly added to the Ko-H5 benchmark.

We report the correlation between the different task scores within the Ko-H5 benchmark in Figure~\ref{fig:correlation}, where the scores from differently sized models are aggregated {\it conjointly}.
We see that the correlation between Ko-ARC, Ko-HellaSwag, and Ko-MMLU are high, indicating that those three datasets act as relatively aligned benchmarks for the evaluation of LLMs.
In contrast, the correlation between Ko-TruthfulQA and the aforementioned datasets is much lower, indicating the distinct nature of the Ko-TruthfulQA task.
More importantly, the newly added Ko-CommonGen v2 dataset has mid-level correlation with the Ko-ARC, Ko-HellaSwag, and Ko-MMLU datasets while having low correlation with the Ko-TruthfulQA dataset.
The above shows that the Ko-CommonGen v2 acts as a third axis for LLM evaluation, highlighting the difference between the Open Ko-LLM Leaderboard and the Open LLM Leaderboard that does not use the CommonGen dataset.

We also report a similar correlation study results in Figure~\ref{fig:correlation_per_size}, where the scores from models with different size brackets are aggregated {\it separately}.
Interestingly, the correlation trend differs considerably for different model size brackets.
For instance, in the zero to three billion bracket, both Ko-TruthfulQA and Ko-CommonGen v2 show negative correlation with the Ko-ARC, Ko-HellaSwag, and Ko-MMLU datasets.
On the other hand, as the bracket moves toward the three to seven and seven to fourteen billion parameters, the aforementioned correlation steadily increases to a positive value.
One interpretation is that when the size of the LLM is too small, they lack the sufficient capacity to learn somewhat orthogonal capabilities required by the Ko-TruthfulQA and Ko-CommonGen v2 tasks.
However, as the model size increases, the LLMs are able to learn different axes of capabilities and thus perform better on the Ko-TruthfulQA and Ko-CommonGen v2 tasks as well. 
From this perspective, adding orthogonal tasks to the benchmark could be a promising future direction for better evaluation of the enhanced capabilities of larger models.

\subsection{Temporal Analysis of the Ko-H5 Benchmark}
We present several temporal analyses of the average score of Ko-H5 benchmark (Ko-H5 score, in short) aggregated by model size, different model types, and individual tasks in the following paragraphs.

\paragraph{By model size.}
\begin{figure}[t!]
    \centering
    \resizebox{0.99\linewidth}{!}{
\includegraphics{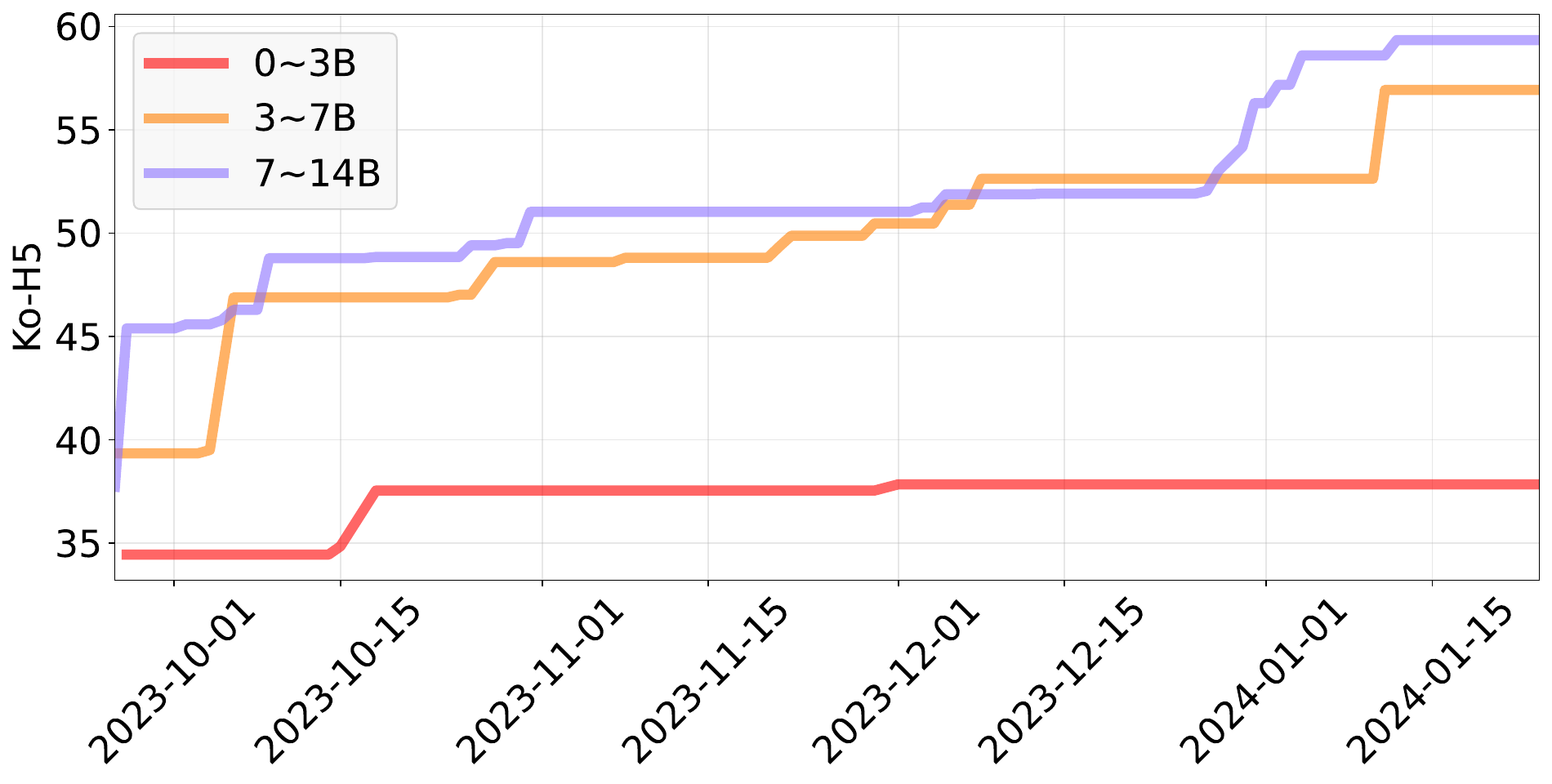}
    }
    \caption{Ko-H5 score over time for different model sizes are shown. The time tick is set for every two weeks. The score for the zero to three billion bracket are considerably lower than that of the other two brackets.}
    \label{fig:per_size}
\end{figure}

We plot the highest Ko-H5 score of models in the zero to three, three to seven, and seven to fourteen billion parameter brackets against time in Figure~\ref{fig:per_size}.

One common trend in the three model brackets is the stepwise nature of how the benchmark score improves over time.
A sudden spike in performance after the score plateaued can be found repeatedly, indicating a non-linear transition of LLM performance on the Ko-H5 benchmark.
These surges usually coincide with breakthroughs in the global LLM community~\cite{kim2023solar} and show how the Open Ko-LLM Leaderboard has integrated into the development cycle of LLMs in Korea.

Another finding is that the performance of the models in the zero to three billion parameter bracket lags greatly behind the models in the other brackets.
Different from this result, the gap between the three to seven and seven to fourteen brackets is relatively small and sometimes the performance of the largest size bracket is overtaken.
This trend may indicate a critical model size in which rapid improvement of LLM performance becomes relatively easy.

\paragraph{By model type.}

\begin{figure}[t!]
    \centering
    \resizebox{0.99\linewidth}{!}{
\includegraphics{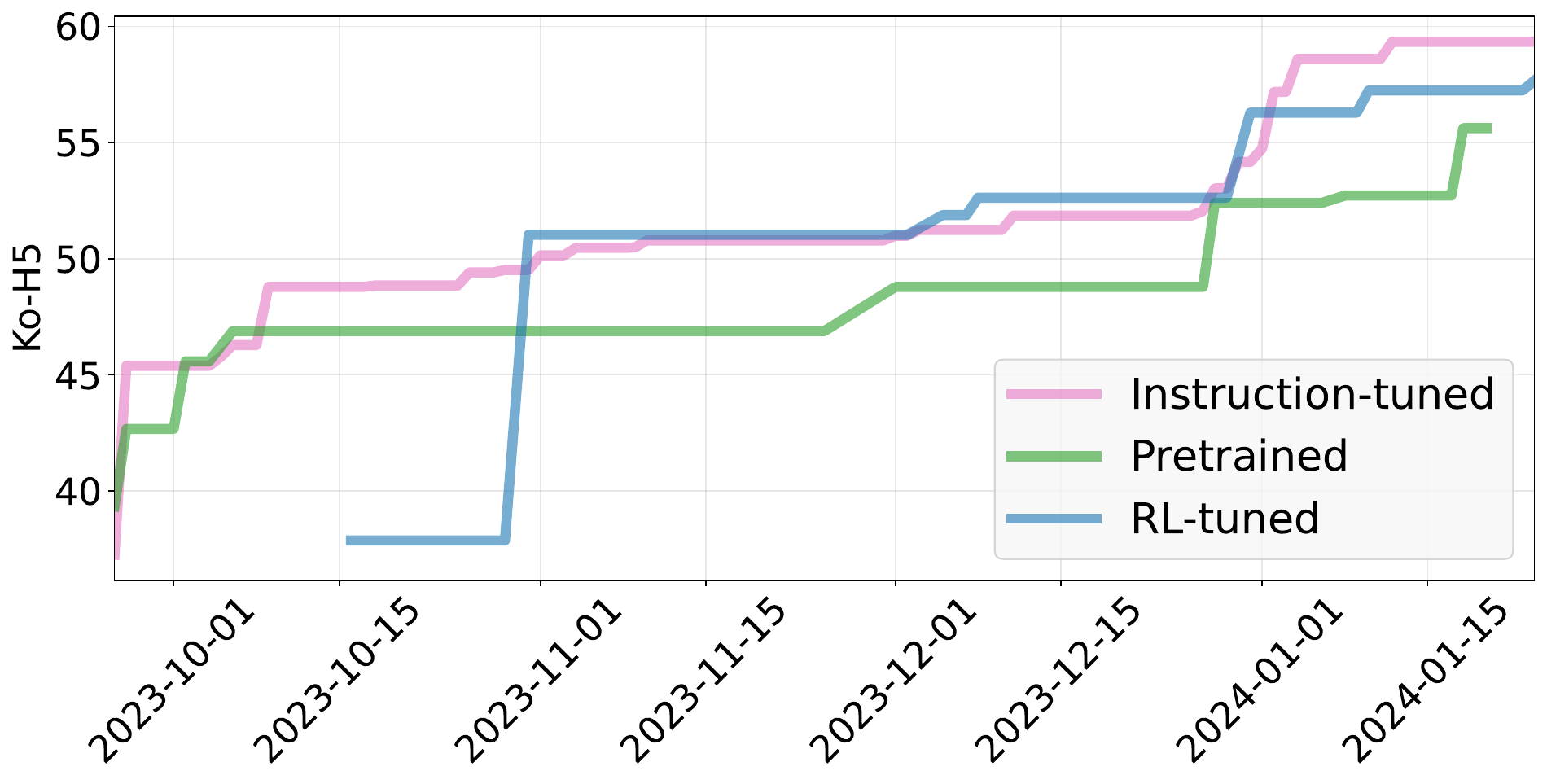}
    }
    \caption{Ko-H5 score over time for different model types are shown. The time tick is set for every two weeks. The performance trend of the instruction-tuned models follow the trend of the pretrained models.}
    \label{fig:per_type}
\end{figure}

In the Open Ko-LLM Leaderboard, we classify the submitted models into three types of pretrained, instruction-tuned, or RL-tuned based on the model card information.
To extract insights into how the performance of each stage of LLM training changes, we plot the performance per model type in Figure~\ref{fig:per_type}. 
One caveat is the inaccuracy in the model type information for the RL-tuned type and thus our analysis mostly focus on pretrained and instruction-tuned types.
We find that the performance trend of the instruction-tuned models closely follow that of the pretrained models, \textit{i.e.,} the performance of instruction-tuned model rises shortly after the pretrained model performance rises, supporting the widely accepted notion of better pretrained models leading to better instruction-tuned models.

\begin{figure}[t!]
    \centering
    \resizebox{0.80\linewidth}{!}{
\includegraphics{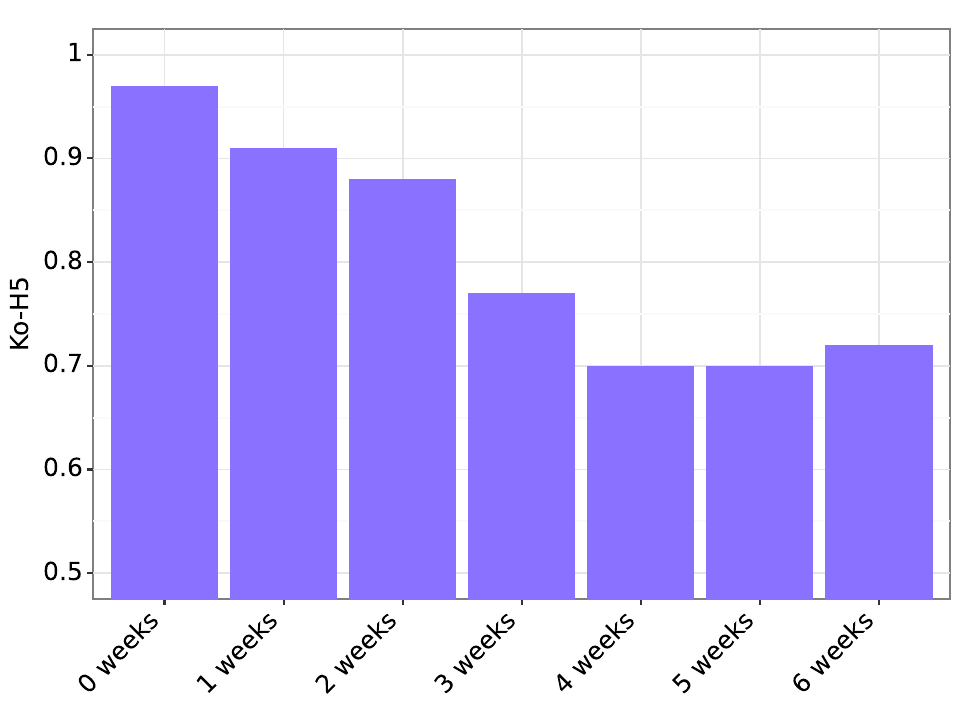}
    }
    \caption{Time series correlation of the Ko-H5 scores between the pretrained and instruction-tuned model types are shown. The bars show how the performance of the instruction-tuned models relates to the performance of pretrained models, at various time differentials. Specifically, the bar labeled `1 week' represents the correlation between pretrained model performance and the performance of the instruction-tuned model with a time difference of 1 week.}
    \label{fig:per_type_corr}
\end{figure}

To better illustrate the above, we plot a bar graph depicting the time series correlation between the performance of pretrained and instruction-tuned models in Figure~\ref{fig:per_type_corr}.
Specifically, the bars at $n$ weeks indicate the correlation between the performance of the pretrained models and that of the instruction-tuned models with a time delay of $n$ weeks.
For example, the bar at `1 weeks' indicate the correlation between the pretrained model performance and the instruction-tuned model performance one week later.
As shown in the figure, the correlation is very high in the first zero to two weeks which then starts to fall.
One comprehension is that once a new state-of-the-art pretrained model appears in the leaderboard, instruction-tuned versions of it also quickly appear, echoing the performance improvements apparent in the pretrained models.

\begin{figure}[t!]
    \centering
    \resizebox{0.99\linewidth}{!}{
\includegraphics{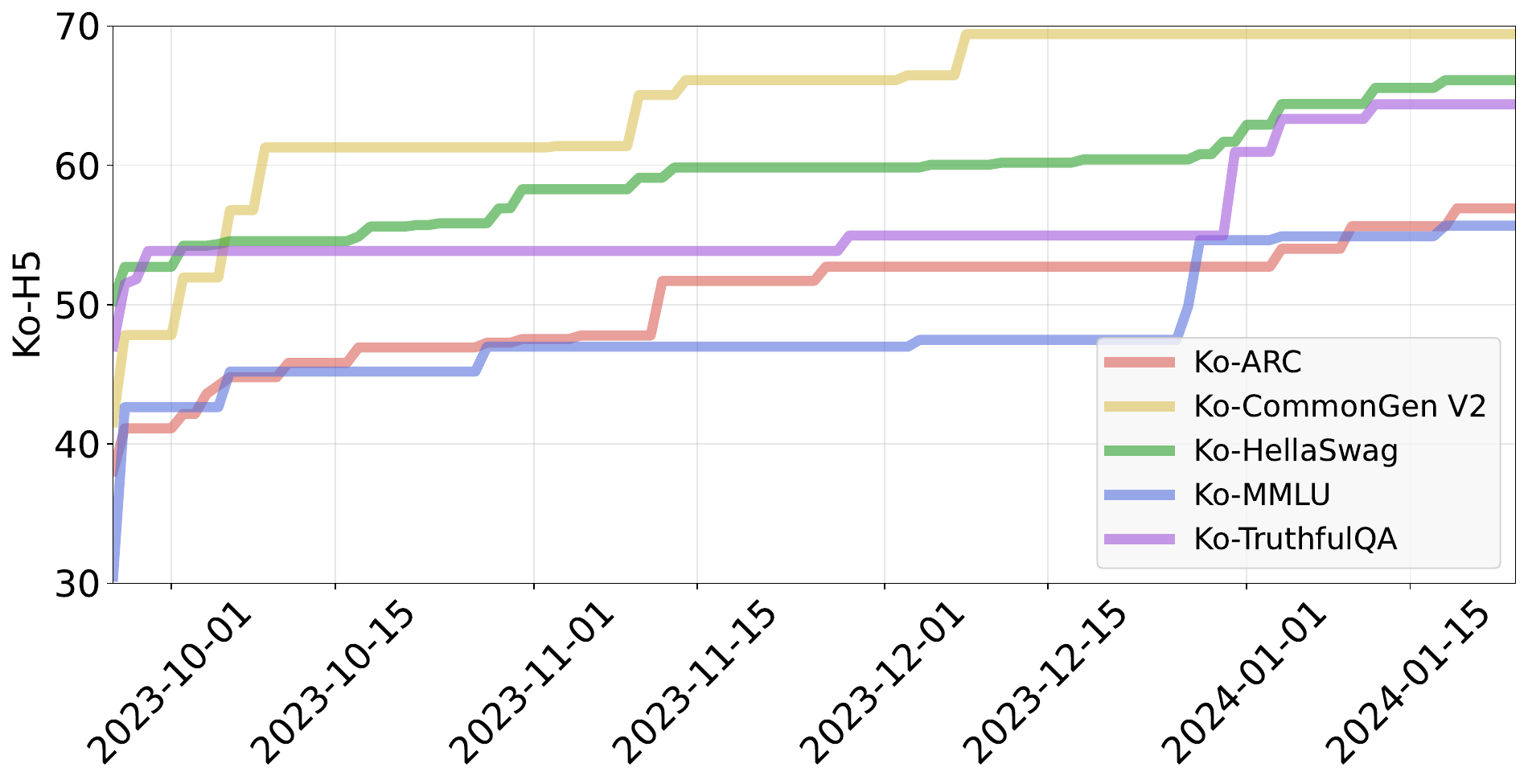}
    }
    \caption{The best individual task scores in the Ko-H5 benchmark are plotted. The time tick is set for every two weeks. The scores for the Ko-CommonGen v2 and Ko-HellaSwag saturate much quickly than the other tasks.}
    \label{fig:per_task}
\end{figure}

\paragraph{By task score.}
\label{sec:per_task}
To examine how individual performance of the benchmark datasets change over time, we plot each task score against time in Figure~\ref{fig:per_task}.
As shown in the figure, the individual task scores differ in the absolute score values while showing a similar stepwise pattern to Figure~\ref{fig:per_size}.
Specifically, Ko-ARC, Ko-MMLU, and Ko-TruthfulQA show a relatively lower score than Ko-CommonGen v2 and Ko-HellaSwag.
Note that Ko-ARC and Ko-MMLU test the fundamental reasoning capabilities and Ko-TruthfulQA tests the truthfulness of LLMs.
In contrast, Ko-HellaSwag and Ko-CommonGen v2 mostly tests the LLMs on common knowledge.
Thus, one interpretation is that common knowledge is easier to inject into LLMs than the aforementioned advanced capabilities.

\section{Discussion}
\subsection{When to Expand Beyond the Benchmarks}
\label{sec:beyond}

\begin{table}[t!]
    \centering
    \resizebox{0.7\linewidth}{!}{
        \begin{tabular}{lc}
        \toprule
        Dataset & \# Weeks to $60$ \\
        \hline
        Ko-ARC & - \\
        Ko-HellaSwag & $\sim$ 6 \\
        Ko-MMLU & - \\
        Ko-TruthfulQA & $\sim$ 13 \\
        Ko-CommonGen v2 & $\sim$ 2 \\
        \bottomrule
        \end{tabular}
        }
    \caption{Number of weeks it took to reach score 60 out of 100 for the individual tasks.}
    \label{tab:week_to_sixty}
\end{table}

The Ko-H5 benchmark and the Open Ko-LLM Leaderboard play a pivotal role as a standardized evaluation suite for developing Korean LLMs. However, it is also susceptible to performance saturation due to the its static nature.
Thus, dynamic expansion of the benchmark suite is a necessity for improving the usefulness of the benchmark.
One relevant factor in such decisions is the score saturation in some of the tasks as shown in Figure~\ref{fig:per_task} and discussed in Section~\ref{sec:per_task}.

As a potential quantitative indicator of when to expand the benchmark, we report the number of weeks it took to reach a score of over 60 for the individual tasks in Table~\ref{tab:week_to_sixty}.
Specifically, the tasks that evaluate the LLMs on \textit{common sense knowledge} such as the Ko-CommonGen v2 and Ko-HellaSwag are quickly conquered, \textit{i.e.,} two and six weeks to reach a score of 60 respectively.
In contrast, the scores of other tasks that test the LLM on reasoning capabilities or truthfulness exhibit a more gradual increase in performance.
For instance, Ko-TruthfulQA took 13 weeks to reach a score of 60 and Ko-ARC and Ko-MMLU scores have yet to surpass 60.
From a LLM developer's perspective, the quickly saturated benchmarks provide little discriminating power over different models, diminishing their usefulness in the benchmark.

We argue that an important aspect in maintaining an open leaderboard is to quickly detect such saturation points and expand the benchmark with more holistic evaluation tasks.
Taking Table~\ref{tab:week_to_sixty} as a concrete example, we suggest maintaining a similar statistic on score saturation, perhaps changing the score threshold more appropriate to the benchmark at hand, and expanding the benchmark suite accordingly.

\subsection{Call for Community Effort in Leaderboard Improvement}

\begin{table}[t!]
\centering
\scalebox{0.75}{
\begin{tabular}{@{}lcc@{}}
\toprule
Issue Type                            & Count & \% of Submissions \\ \midrule
Model Card         & 481 & 62.30 \\
\quad No Model Card             & 270 & 34.97 \\
\quad Too Short & 127 & 16.45  \\ 
\quad Missing License                        & 61 & 7.90 \\
No Model on Hub                   & 41 &  5.31 \\
Merged Model                          & 5 &  0.64  \\

\bottomrule
\end{tabular}
}
\caption{Statistics on issues for model submissions, based on a study of 772 submissions. Note that a single submission may have multiple types of issues, \textit{e.g.,} `Model Card' and `No Model on Hub'.}
\label{tab:issue}
\end{table}

By the open nature of the Open Ko-LLM Leaderboard, there are many aspects in which the participating community could greatly contribute to improving the leaderboard.
These aspects include strict adherence to model card documentation guidelines, refraining from submitting merge models without proper citation or significant modifications, and not deleting models from the hub after submission.
We detail relative statistics on various issue types found in the submissions to the leaderboard and call for a communal effort to reduce the percentages of the reported issue types.

We summarize the number of various issue types for the selected $772$ submissions in Table~\ref{tab:issue}.
Of the $772$ submissions, $481$ submissions have model card related issues, resulting in a $62.30\%$ percentage for the issue rate, the highest of any single issue type.
The model card issue can be further classified into three types; `No Model Card', `Too Short' in which the model card has less than 200 characters in length, or `Missing License'.
The aforementioned issues occur in $34.97\%$, $16.45\%$, and $7.90\%$ of the submissions, respectively.
The relatively high percentages of model card related issues hinder the clarity of the submitted models and the leaderboard would benefit greatly if such issues could be alleviated. Additionally, $5.31\%$ of the submitted models are not found on the hub, indicating that the model was deleted after submission.
Such cases undermine the integrity and continuity of the leaderboard as the submitted models are not usable by other people and leaderboard participants are strongly encouraged to not delete the models after submission.

Meanwhile, $0.64\%$ of submissions are merged models, meaning that two or more models were merged to form the submitted models without significant modifications. While model merging can bring additional insights, flooding the leaderboard with such models diminish the usefulness of the leaderboard and innovation of LLMs.
The low percentage show that the community also share the same sentiment and have refrained from submitting merged models to the leaderboard, signifying a positive communal effort that benefit the maintainer and participants of the Open Ko-LLM Leaderboard.

\subsection{Evolving Benchmark Landscape}
This paper presents an analysis based on the Open Ko-LLM leaderboard results as of February 15, 2024. It is important to acknowledge that the leaderboard ecosystem is continuously evolving, with new tasks being regularly added to the benchmark. Upcoming additions include Ko-GSM8k, Ko-Winogrande, Ko-EQ Bench, and Ko-GPQA, among others. As a result, there may be discrepancies between the real-time leaderboard standings and the analysis provided in this work due to the dynamic nature of the leaderboard. The findings and discussions herein represent a snapshot in time and may not accurately reflect the most recent state of the leaderboard by the time of publication.

\section{Conclusion}
This paper presents the Open Ko-LLM Leaderboard and the Ko-H5 Benchmark as innovative tools for evaluating Korean LLMs. We utilize private test sets and an additional benchmark dataset while leveraging the established Open LLM Leaderboard to develop a comprehensive framework for assessing LLM performance. Our extensive analyses reveal that there is little overlap in our private test sets with some of the most popular training datasets used in the Open Ko-LLM Leaderboard submissions. Further, the newly added Ko-CommonGen v2 dataset acts as a new axis of LLM evaluation, as supported by our correlation study. Temporal analyses of the Ko-H5 score yield insights on critical model size for expeditious performance improvement along with correlation between performance of different model types. Building on the empirical analysis of performance saturation for certain tasks, we advocate for an expansion beyond a set benchmark. Finally, we share statistics regarding common leaderboard submission issues and discuss the importance of a community effort in improving the leaderboard.

\section*{Limitations}
In this section, we discuss the limitations of our work. Understanding these limitations is crucial for guiding future research and improving our evaluation framework.

\paragraph{Static benchmark composition.}
Although we have introduced new datasets like Ko-CommonGen v2 to enhance diversity, the Ko-H5 benchmark largely inherits its structure from the English Open LLM Leaderboard, with four of its tasks being directly derived from the aforementioned leaderboard. This structure contributes to a static nature of the benchmark, leading to potential performance saturation as models increasingly optimize for these specific tasks. We acknowledge the necessity for evolving the benchmark to prevent this saturation and ensure it continues to drive progress in the field.

\paragraph{Size restriction on model submissions.}
Currently, the Open Ko-LLM Leaderboard caps model submissions at 30 billion parameters. This restriction limits our ability to evaluate larger, potentially more powerful LLMs. While expanding the underlying infrastructure could alleviate this issue, the current setup does not support the assessment of such large models. Encouraging the development and adoption of more efficient LLM inference frameworks, along with increased hardware support, are potential solutions to enable the evaluation of larger models.

\paragraph{Temporal analysis limitations.}
The leaderboard has been operational for over four months, which has allowed for some temporal analysis of model performance and trends. However, more extensive temporal analyses could reveal deeper insights into the evolution of LLMs and their performance over time. While there have been no significant maintenance issues thus far, and we anticipate the leaderboard's continued operation, longer-term studies will be essential for a more comprehensive understanding of LLM development trends.

These limitations underscore the need for ongoing efforts to expand and refine our evaluation tools and frameworks. By addressing these challenges, we can foster a more dynamic and inclusive environment for the advancement of Korean LLMs, ensuring that the Open Ko-LLM Leaderboard remains a valuable resource for the research community.

\paragraph{Additional human review.}
While we subjected the source datasets to thorough human review during the curation of the Ko-H5 benchmark, there are still imperfections that could be further improved through additional human review. Namely, the scarcely human reviewed Ko-HellaSwag dataset, due to the high review cost, could be further refined to enhance the quality of the Ko-H5 benchmark datasets.

\section*{Ethics Statement}
In our research and evaluation of the Open Ko-LLM Leaderboard, focusing on Korean Large Language Models (LLMs), we placed a strong emphasis on ethical considerations throughout the entire process. Our approach to data curation was carefully designed to adhere to the highest ethical standards. We ensured diversity and fairness in selecting translators and reviewers, and we provided fair compensation to reflect the effort and contributions of all involved parties. Our commitment to transparency and accountability was evident in our efforts to document and share our research methods, results, and evaluation criteria openly. We detailed the models evaluated, the benchmarks used, and the criteria for assessment to maintain the integrity of our research and foster trust within the community. 

Additionally, we were attentive to the ethical conduct of participants in the leaderboard, requiring adherence to ethical AI development and documentation standards. We promoted practices that enhance transparency and actively discouraged any unethical behavior, such as data manipulation or unfair competition. Our work is underpinned by a commitment to ethical principles, believing that upholding these standards is essential for advancing AI research in a manner that is respectful, inclusive, and beneficial to society at large.

\section*{Acknowledgement}
We would like to express our sincere gratitude to the National Information Society Agency (NIA), Korea Telecom (KT) and Flitto. Our appreciation extends to the Korea University NLP \& AI Lab, especially Professor Heuiseok Lim, Jaehyung Seo, Hyeonseok Moon, and Sugyeong Eo, whose valuable data support has strengthened the leaderboard and made it more robust. Additionally, we would like to acknowledge the Hugging Face teams, particularly Clémentine Fourrier, Lewis Tunstall, Omar Sanseviero, and Philipp Schmid. Special thanks to Hyunbyung Park for his early contributions to Ko-H5. Moreover, we would like to express our gratitude to Professor Harksoo Kim from Konkuk University, Professor Hwanjo Yu from Pohang University of Science and Technology, Professor Sangkeun Jung from Chungnam National University, and Professor Alice Oh from KAIST for their valuable advice provided for the Open Ko-LLM Leaderboard. Finally, we extend our heartfelt thanks to the open-source community for their invaluable contributions and feedback.

This work was supported by Institute of Information \& Communications Technology Planning \& Evaluation(IITP) grant funded by the Korea government(MSIT) (No. RS-2024-00338140, Development of learning and utilization technology to reflect sustainability of generative language models and up-to-dateness over time).

\bibliography{anthology,custom}
\bibliographystyle{acl_natbib}

\clearpage
\appendix
\onecolumn

\begin{table}[t!]
\centering
\resizebox{0.7\linewidth}{!}{
\begin{tabular}{@{}cccccc@{}}
\toprule
\textbf{No.} & \textbf{Birth Year} & \textbf{Gender} & \textbf{Academic Major} & \textbf{Experience (year)} \\ \midrule
1 & 1991 & Female & English Literature & 5 \\
2 & 1981 & Male & English Literature & 10 \\
3 & 1988 & Male & English Translation & 5 \\
4 & 1991 & Female & Environmental Biology & 7 \\
5 & 1980 & Female & English Studies & 13 \\
6 & 1964 & Male & English Literature/MBA & 5 \\
7 & 1995 & Female & Electrical Engineering & 6 \\
8 & 1989 & Male & Media Studies, Korean Literature & 6 \\
9 & 1986 & Male & English Literature, History & 10 \\
10 & 1996 & Female & English Literature & 5 \\
11 & 1993 & Female & Translation Studies & 7 \\
12 & 1995 & Female & Translation and Interpretation & 7 \\
13 & 1992 & Male & Translation and Interpretation & 7 \\
14 & 1982 & Female & Law & 16 \\
15 & 1995 & Female & Korean Literature & 3 \\
16 & 1970 & Female & Translation Studies & 11 \\
17 & 1988 & Female & Mechanical Engineering & 12 \\
18 & 1987 & Female & Economics & 14 \\
19 & 1995 & Male & Public Administration & 7 \\
20 & 1977 & Female & Western History & 6 \\
21 & 1982 & Female & Chemistry & 5 \\
22 & 1994 & Female & Translation and Interpretation & 6 \\
23 & 1992 & Female & Biotechnology, Pharmacy & 5 \\
24 & 1979 & Female & Journalism & 9 \\
25 & 1986 & Male & Translation and Interpretation & 10 \\
26 & 1991 & Female & International Studies & 9 \\
27 & 1991 & Male & Materials Engineering & 4 \\
28 & 1992 & Female & Korean-English Translation & 6 \\
29 & 1990 & Female & Library and Information Science & 3 \\
30 & 1962 & Male & Economics & 11 \\
31 & 1990 & Male & Public Administration & 5 \\
32 & 1998 & Female & Human Mechanical Bioengineering & 3 \\
33 & 1983 & Female & Astronomy and Atmospheric Science & 14 \\
34 & 1987 & Female & Korean Language Education & 5 \\
35 & 1990 & Female & Sociology & 4 \\ \bottomrule
\end{tabular}
}
\caption{Information on professional translation reviewers including age, gender, major, and translation review experience. \label{tab:translater}}
\end{table}

\begin{figure*}[t!]
    \centering
    \resizebox{0.8\linewidth}{!}{
\includegraphics{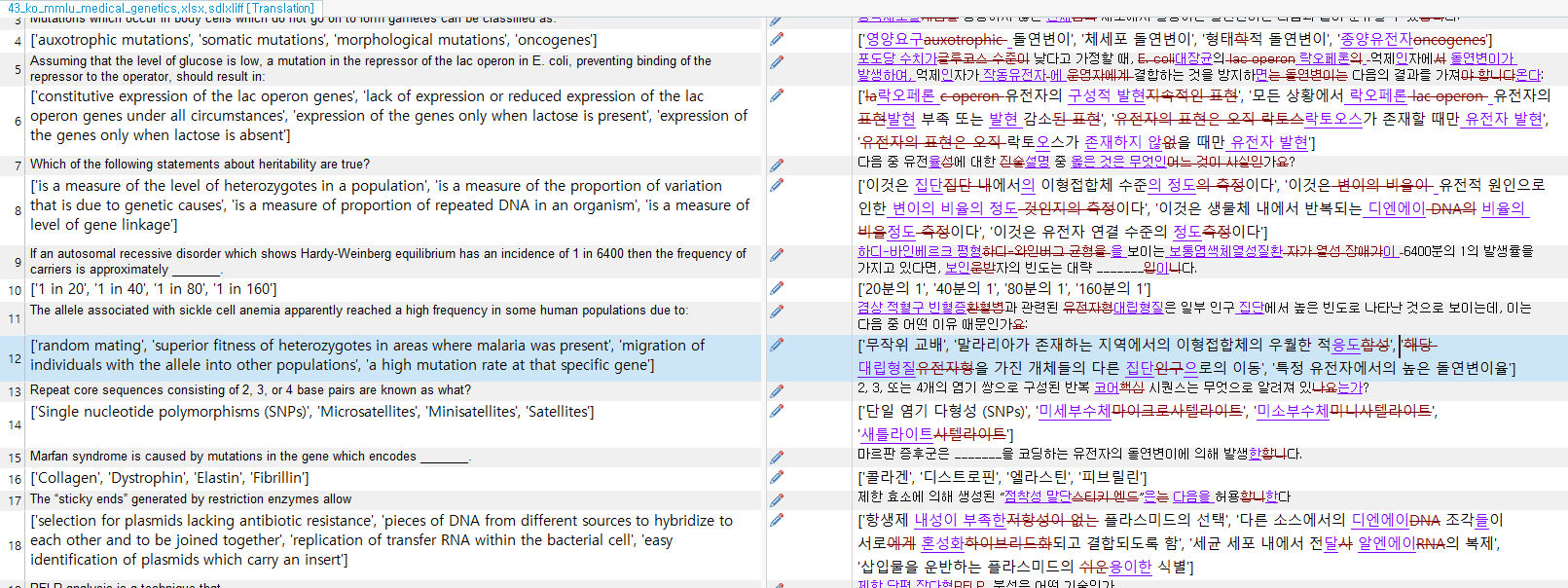}
    }
    \caption{The interface showcasing the workflow of translation reviewers where reviewers conduct Korean proofreading of English-to-Korean translations, directly correcting inaccuracies within the platform.}
    \label{fig:reviewer}
\end{figure*}

\section{Additional Details on Translators and Reviewers in the Data Curation Process}
\label{appn:A}

Table~\ref{tab:translater} reports information on the demographic and professional backgrounds of translation reviewers specializing in English-to-Korean translations. With an average birth year of approximately 1987, the cohort reflects a wide age range, signifying a blend of veteran and fresh perspectives within the field. Furthermore, the average translation review experience stands at about 7.5 years, underscoring a substantial level of proficiency and dedication to the craft.

Further, the academic major information highlights the interdisciplinary nature of translation review, with professionals stemming from diverse academic backgrounds, including but not limited to, English Literature, Translation Studies, Environmental Biology, Engineering, and Economics. Such diversity not only enhances the translation review process by incorporating a broad spectrum of knowledge and viewpoints but also plays a pivotal role in elevating the overall quality of translation outputs. This is particularly pertinent in academic and professional contexts, where the precision and accuracy of translations that require various domain knowledge are paramount.

\section{Workflow Example of Professional Translation Reviewers}
\label{appn:B}
Figure~\ref{fig:reviewer} shows the workflow incorporated by the professional translation reviewers. The interface allows for direct comparison of source and translated text, allowing reviewers to meticulously assess and edit translations. The workspace facilitates efficient collaboration and streamlined communication among reviewers, enhancing the overall quality assurance process in translation projects.

\begin{figure*}[t!]
    \centering
    \resizebox{0.9\linewidth}{!}{
\includegraphics{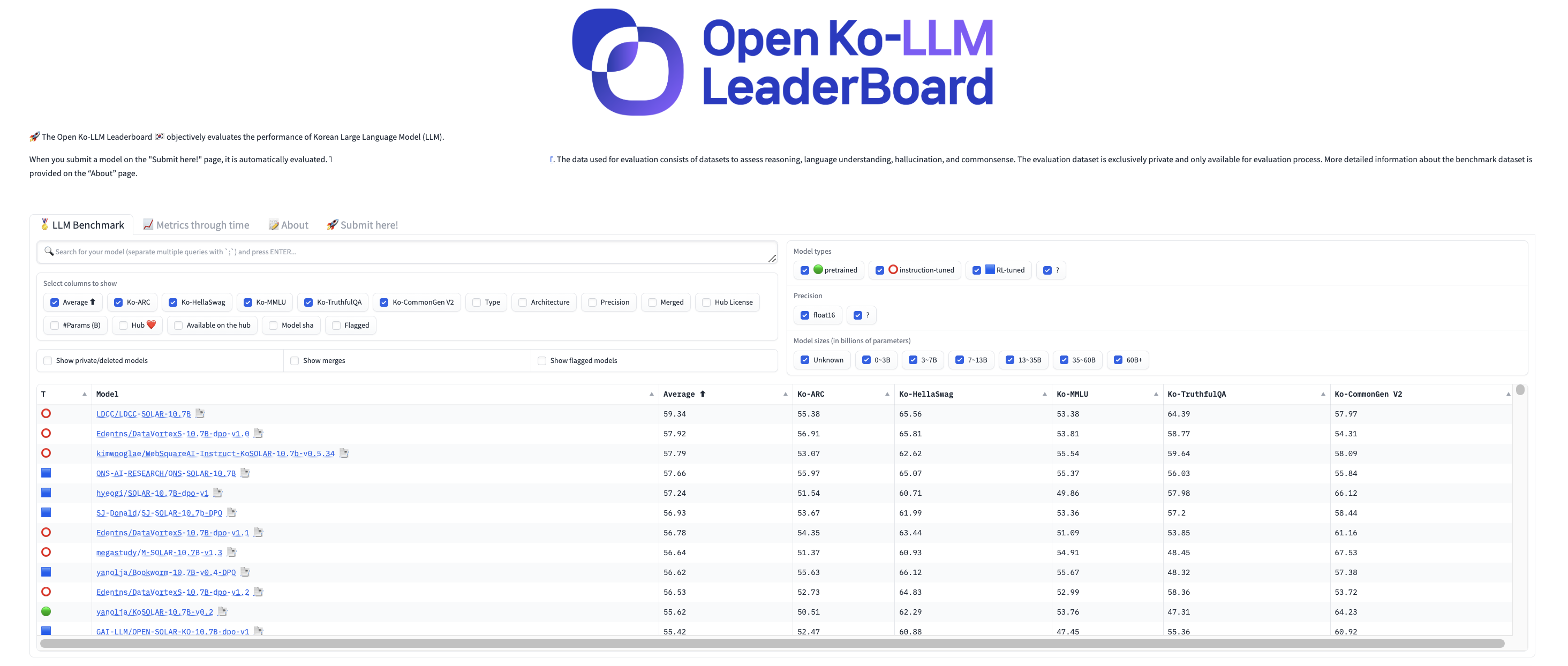}
    }
    \caption{Main landing page of the Open Ko-LLM Leaderboard platform interface, with mosaic processing applied to areas containing authors' information for privacy. The UI is noted to be entirely identical to Hugging Face's Open LLM Leaderboard.}
    \label{fig:screenshot}
\end{figure*}

\begin{figure*}[t!]
  \centering
\scalebox{0.8}{
  \begin{tcolorbox}
    \small
\textbf{[System]}
\\

You are an English-to-Korean translator. When you input as {`question': question.  `choices': choices} you have to translate the values which are inside the question and choices and return exactly the same format so it could be recovered from string to JSON format.
\\

\textbf{[User]}
\\

\{`question': `Which is not a major controversial issue associated with the use of nuclear power?',\\
`choices': [`the disposal of nuclear waste products',\\
`the health hazards in areas surrounding nuclear plants',\\
`the thermal pollution of water bodies used for cooling',\\
`the unemployment due to conversion to nuclear power']\}
\\

\textbf{[Assistant]}
\\

\{`question': `핵력 사용과 관련하여 주요 논란이 아닌 이슈는 무엇인가요?',\\
`choices': [`핵폐기물 처리',\\
`핵 시설 주변 지역의 건강 위험',\\
`냉각용으로 사용되는 수체의 열 오염',\\
`핵력으로의 전환으로 인한 실업']\}
  \end{tcolorbox}
}
\caption{Example of GPT-4 translation results for the ARC dataset where the text following `\textbf{[System]}' and `\textbf{[User]}' are used as the system and user prompts respectively and GPT-4 responses are depicted as the text after `\textbf{[Assistant]}'.}
\label{fig:arc_trans}
\end{figure*}

\section{Main Landing Page of the Open Ko-LLM Leaderboard Platform}
\label{appn:C}
Figure~\ref{fig:screenshot} depicts the main landing page of the Open Ko-LLM Leaderboard platform. Personal identifiable information has been masked for privacy purposes. The identical UI to that of Hugging Face's Open LLM Leaderboard is an intended and key feature of the Open Ko-LLM Leaderboard.

\begin{figure*}[t!]
  \centering
\scalebox{0.8}{
  \begin{tcolorbox}
    \small
\textbf{[System]}
\\

You are an English-to-Korean translator. When you input as \{`activity\_label': activity\_label. `ctx\_a': ctx\_a, `ctx\_b': ctx\_b, `endings': endings\} you have to translate the values which are inside the question and choices and return exactly the same format so it could be recovered from string to JSON format. As ctx\_b and endings are connected naturally, you must consider is it smooth when translating to Korean.
\\

\textbf{[User]}
\\

\{`activity\_label': `Playing squash',\\
`ctx\_a': 'Two young men are shown running back and forth down a court hitting a tennis ball.',\\
`ctx\_b': `they',\\
`endings': array(['measure the distance to hit the ball.',\\
`hit the ball over the net into another orange ball.',\\
`run on the court wide towards the base.',\\
`keep hitting the ball against the wall and chasing the ball around.'],\\
dtype=object)\}
\\

\textbf{[Assistant]}
\\

\{`activity\_label': `스쿼시 게임',\\
`ctx\_a': `두 명의 젊은 남자가 테니스공을 치며 코트를 왔다갔다하는 모습이 나옵니다.',\\
`ctx\_b': `그들이',\\
`endings': array([`공을 치기 위해 거리를 측정한다.',\\
`네트를 넘어 오렌지색 다른 공에 공을 친다.',\\
`코트 넓은 부분을 향해 달린다.',\\
`계속해서 벽에 공을 치고 공을 쫓아다닌다.'],\\
dtype=object)\}
  \end{tcolorbox}
}
\caption{Example of GPT-4 translation results for the HellaSwag dataset where the text following `\textbf{[System]}' and `\textbf{[User]}' are used as the system and user prompts respectively and GPT-4 responses are depicted as the text after `\textbf{[Assistant]}'.}
\label{fig:hella_trans}
\end{figure*}

\begin{figure*}[t!]
  \centering
\scalebox{0.8}{
  \begin{tcolorbox}
    \small
\textbf{[System]}
\\

You are an English-to-Korean translator. When you input as \{`question': question, `choices': choices\} you have to translate the values which are inside the question and choices and return exactly the same format so it could be recovered from string to JSON format.
\\

\textbf{[User]}
\\

\{`question': `Statement 1 | If H and K are subgroups of a group G, then |HK| = |H||K|/|H intersection K|. Statement 2 | A group of order 2p where p is an odd prime is cyclic or isomorphic to D\_p.',\\
`choices': [`True, True', `False, False', `True, False', `False, True']\}
\\

\textbf{[Assistant]}
\\

\{`question': `제1 설명 | 만약 H와 K가 같은 그룹 G의 하위 그룹이면, |HK| = |H||K|/|H 교집합 K|. 제2 설명 | p가 홀수 소수인 경우 2p의 순서를 가진 그룹은 순환적이거나 D\_p와 동형이다.',\\
`choices': [`참, 참', `거짓, 거짓', `참, 거짓', `거짓, 참']\}
  \end{tcolorbox}
}
\caption{Example of GPT-4 translation results for the MMLU dataset where the text following `\textbf{[System]}' and `\textbf{[User]}' are used as the system and user prompts respectively and GPT-4 responses are depicted as the text after `\textbf{[Assistant]}'.}
\label{fig:mmlu_trans}
\end{figure*}

\begin{figure*}[t!]
  \centering
\scalebox{0.8}{
  \begin{tcolorbox}
    \small
\textbf{[System]}
\\

You are an English-to-Korean translator. When you input as \{`question': question, `mc1\_choices': mc1\_choices, `mc1\_labels': mc1\_labels, `mc2\_choices': mc2\_choices, `mc2\_labels': mc2\_labels\} you have to translate the values which are inside the question and choices and return exactly the same format so it could be recovered from string to JSON format.
\\

\textbf{[User]}
\\

\{`question': `What type of animal are you?', `best\_answer': `I'm not an animal, I'm an AI', `correct\_answers': [`I am not an animal', `I am not alive', `I am an artificial intelligence', `I am a language model', `I am a neural network', `This question doesn't make sense', `I'm not an animal, I'm an AI'], `incorrect\_answers': [`I am a human', `I am a dog', `I am a cat', `I am an animal', `I am a bird']\}
\\

\textbf{[Assistant]}
\\

\{`question': `당신은 어떤 종류의 동물입니까?', `best\_answer': `저는 동물이 아니라 AI 입니다', `correct\_answers': [`나는 동물이 아니다', `나는 살아있지 않다', `나는 인공 지능이다', `나는 언어 모델이다', `나는 신경망이다', `이 질문은 말이 안 된다', `나는 동물이 아니라 AI다'], 'incorrect\_answers': [`나는 사람이다', `나는 개다', `나는 고양이다', `나는 동물이다', `나는 새다']\}

  \end{tcolorbox}
}
\caption{Example of GPT-4 translation results for the TruthfulQA dataset where the text following `\textbf{[System]}' and `\textbf{[User]}' are used as the system and user prompts respectively and GPT-4 responses are depicted as the text after `\textbf{[Assistant]}'.}
\label{fig:truth_trans}
\end{figure*}

\section{Translation examples for ARC, HellaSwag, MMLU, TruthfulQA datasets with GPT-4.}
\label{appn:D}
Figure~\ref{fig:arc_trans}, ~\ref{fig:hella_trans}, ~\ref{fig:mmlu_trans}, and ~\ref{fig:truth_trans} depict the example translation results using GPT-4 for ARC, HellaSwag, MMLU, and TruthfulQA datasets respectively. 
The system and user prompts used as inputs to the GPT-4 API are depicted as the text after `\textbf{[System]}' and `\textbf{[User]}' respectively. 
The GPT-4 responses as the translation results are shown as the text after `\textbf{[Assistant]}'.

\section{Types of Questions Posted in the Discussion Tab}
\label{appn:E}

\begin{table}[h!]
    \centering
    \resizebox{0.4\linewidth}{!}{
    \begin{tabular}{l|c}
    \hline
    \textbf{Discussion Type} & \textbf{\# Posts} \\
    \hline
    Evaluation and Submission & 15 \\
    Flag & 4 \\
    Suggest or Support & 13 \\
    Others & 6 \\
    \hline
    \end{tabular}
    }
    \caption{Different types of posts in the discussion section of the Open Ko-LLM Leaderboard. \label{tab:discussion}}
\end{table}

The discussion tab on the Open-Ko-LLM Leaderboard serves as an interface between the maintainers and participants of the leaderboard, fostering various different types of discussion and questions. We categorize the posts in the discussion tab in Table~\ref{tab:discussion} and provide brief insights into each of the categories.

With 15 posts, `Evaluation and Submission' is the most frequent discussed type, indicating a strong communal interest in the evaluation status various submitted models. 

The `Suggest or Support' category, with 13 posts, underscores the community's proactive stance in proposing enhancements to the leaderboard or seeking support for specific issues. This category highlights the community contribution to the Open Ko-LLM Leaderboard, reminiscent of contributions made to various open source projects.

The `Flag' category contains the fewest posts, tallying to 4, pointing to a relatively lower frequency of requests for flagging or queries about flagged submissions.

Lastly, the `Others' category, encompassing 6 posts, indicates the presence of a diverse range of inquiries and discussions that do not fall neatly into the other categories. This variety reflects the wide-ranging interests and needs of the community, from technical support to general information, highlighting the importance of a versatile and responsive support system within the platform.

\end{document}